# A More Efficient Chinese Named Entity Recognition base on BERT and Syntactic Analysis

Xiao Fu[*], Guijun Zhang

Zhejiang Informatization Development Institute, Hangzhou Dianzi University, Hangzhou, 310018, P.R. China;

*Corresponding author: fuxiao@hdu.edu.cn

**Abstract**

We propose a new Named entity recognition (NER) method to effectively make use of the results of Part-of-speech (POS) tagging, Chinese word segmentation (CWS) and parsing while avoiding NER error caused by POS tagging error. This paper first uses Stanford natural language process (NLP) tool to annotate large-scale untagged data so as to reduce the dependence on the tagged data; then a new NLP model, g-BERT model, is designed to compress Bidirectional Encoder Representations from Transformers (BERT) model in order to reduce calculation quantity; finally, the model is evaluated based on Chinese NER dataset. The experimental results show that the calculation quantity in g-BERT model is reduced by 60% and performance improves by 2% with Test F1 to 96.5 compared with that in BERT model.

## 1 Introduction

Named entity recognition (NER) aims to identify entity names in the text and classify them into different categories such as person, location, organization, etc. NER is an important task in processing natural language and it is an essential for many downstream applications such as entity linking, event extraction and question answering.

Although significant progress has been made by these methods in Chinese NER task, there are still some issues remain unsolved. First, a significant flaw is that only a very small amount of labeled data is available. Second, how to effectively integrate CWS, POS tagging and parsing into NER remains a question. Next, it can effectively avoid NER error caused by error in upstream tasks as well as the decline in algorithm efficiency caused by serial task. Finally, the current optimal NER model [1] contains 110M parameters, causing high calculation cost.

In the process of natural language, each individual word in English can represent a complete semantic unit while only a phrase can express a complete meaning in Chinese, so it is important to consider the difference between Chinese and English in the process. Devlin et al. [1] indicated that MLM [2] significantly improved BERT model performance, but its model focused mainly on the original language signal and less on semantic knowledge unit modeling. For example, when BERT model is used to process Chinese, it is difficult to learn the complete semantic meaning of a larger semantic unit by predicting Chinese characters to build the model. For example, words like 俄[mask]斯, [mask]工智能 can be easily inferred by BERT model based on word collocation. But there is no explicit model to semantic unit (such as Russia, artificial intelligence) and the corresponding semantic relationship. The language model for Chinese needs to be established on the understanding of words and it is important to avoid the model only learns partial word collocation.

In the basic process of natural Chinese, information including CWS, POS tagging and parsing can boost the effect of NER algorithm [3-5]. For example, Mintz et al. [6] used Stanford NLP model to label existing data. Comparing with weak supervision learning of adjacent sentences, labeling information such as CWS or POS tagging promotes the model's understanding of sentences. Although it has been proven that large scale corpora can automatically master features such as CWS and POS tagging [7], supervision data expedite the model training. The Distillation method [8] compressed the trained CWS, POS tagging and parsing model in large-scale corpus into a pre-training model before using NER label data for fine tunes. This method indirectly realizes the target of using information such as POS tagging to label, meanwhile, it gets rid of NER direct dependence on upstream POS tagging so as to avoid NER error caused by POS tagging error.

In this paper, we first use Stanford NLP tool to label CWS、parsing and POS tagging in data in order to create labeled data. Second, inspired by BERT model, we randomly mask some tokens in the input language model in an attempt to predict the original vocabulary id of the masked word based merely on the context. Unlike left-to-right language model pre-training, MLM target allows for the fusion of left and right context representations, which enables us to resort to deep bidirectional converter in pre-training. In addition to MLM, we also introduce tasks such as POS tagging that jointly pre-trains text-pair representations. Then, we compress BERT model by adding gate node. First, a global node is added in each sentence and the node inputs the most important information in each sentence. The modification and delivery of information is determined by a gate node. This node can help remember the most important information in sentence comprehension. Compared with multiple self-attention heads mechanism, this node can be regarded as a fully deployed multiple attention and number of self-attention heads is equal to that of hidden nodes.

Finally, the model is evaluated on the basis of widely used Chinese NER datasets. Experimental results show that g-BERT model proposed by us performs better than other state-of-the-art methods and gains new benchmarks.

The contributions of our paper are as follows:

(1) We add CWS and POS tagging in the pre-training and promote the model's comprehension of sentence by using Stanford nlp tool to label data.

(2) In the representation of the sentence, we add the global node to input the max-pooling in each sentence, and output the modified content of the gate node, which ultimately reduces model structure.

(3) Our g-BERT model reduces calculation quantity by 60% and improves performance by 2%, with Test F1 to 96.5 (2 absolute improvement)

**2 Related Work**

2.1 NER

By assigning a unique label to each sentential word, NER is widely regarded as a sequence labeling problem [9]. Early studies on sequence labeling used such models as HMM, MEMM, and CRF [10]. Recently, neural network models have been successfully applied to sequence labeling [11-13]. In these work, the model uses transformer to extract feature and make fine tunes so as to achieve state-of-the-art performance [1, 14], which is exploited as the baseline model in our work.

2.2 BERT fine-tuning approaches

Language model pre-training has been proven to be effective in improving many natural language processing tasks [15-19]. These tasks include sentence-level tasks, such as natural language deduction [20,21]and paraphrase [22], predicting the relationship between sentences through holistic analysis as well as token-level tasks such as naming entity recognition [23] and answer SQuAD (Stanford Question Answering Dataset) question [24], where models are required to produce fine-grained output at the token-level.

The fine-tuning approach first performs unsupervised pre-training on large-scale Internet corpus and then it is tuned based on the feature. For example, Mikolov et al. [25], Sutskever et al.[26], Le et al. [27] built language model task to train word vectors, which can significantly improve the effect of multiple tasks. GPT utilized transform models to train dynamic words. Then the subsequent different multiple tasks were unified into the single frame for summary, obtaining the highest recorded scores in multiple tasks. Devlin et al. [1] achieved the optimal level at present by introducing MLM and NSP tasks and making full use of context information of the sentences.

It remains a heated research topic over the decades for researches to study widely applicable words representations. In n-gram [25], pretrained word embeddings are considered to be an integral part of modern NLP systems, such as word embedding of unsupervised learning. Arora et al.[28] constructed static sentence vector by using word frequency and word embedding. ELMo [16] generalized traditional word embedding research in a different dimension and aimed to extract context-sensitive features by using a language model in which contextual word embeddings are integrated with existing task-specific architectures.

2.3 Gates

In the process of natural language, the gate mechanism is used to deal with plant distance correlation. Hochreiter et al. [29] used input gates, output gates and forgetting gates to achieve long and short memory of information. Heck and Salem [30] reduced the number of gates in LSTM and simplified calculation complexity in it. Dauphin et al. [31] added linear gate nodes into the language model, demonstrating that using such node is more efficient. Gehring et al. [32] combined convolution and gate nodes to achieve the efficient calculation and long-range dependence of neural networks.

**3. Model**

3.1 Model architecture

g-BERT is a multi-layer bidirectional Transformer encoder model based on the original implementation described by Devlin et al. [1] each layer contains one gate and multiple transformers nodes. We choose gated linear units [31] as non-linearity which implement a simple gating mechanism over the output of the convolution $V([A\ B]) = A \otimes \sigma(B)$, where $A, B \in R^H$, A is the state of upper layer of the gate and B represents all-token node, max-pooling in the feature. $\otimes$ is the point-wise multiplication.

The gate $\sigma(B)$ controls what is relevant in inputting A in the current context. A similar nonlinearity has been introduced in Oord et al. [33] who applied tanh to A but Dauphin et al. [31] showed that GLUs perform better in the context of language modeling.

In this work, we set layer number (i.e., Transformer blocks) as L, the hidden size as H, and the number of self-attention heads as A. In all cases we set the feed-forward/filter size as 4H, i.e., 768 for H =192. BERTBASE: L=12, H=192, A=4, Total Parameters=30M.

BERTBASE is chosen in a model size as L=12, H=768, A=12, Total Parameters=110M. Model size is four times as large as ours.

The comparison between g-BERT, and BERT is visually shown in Fig 1.

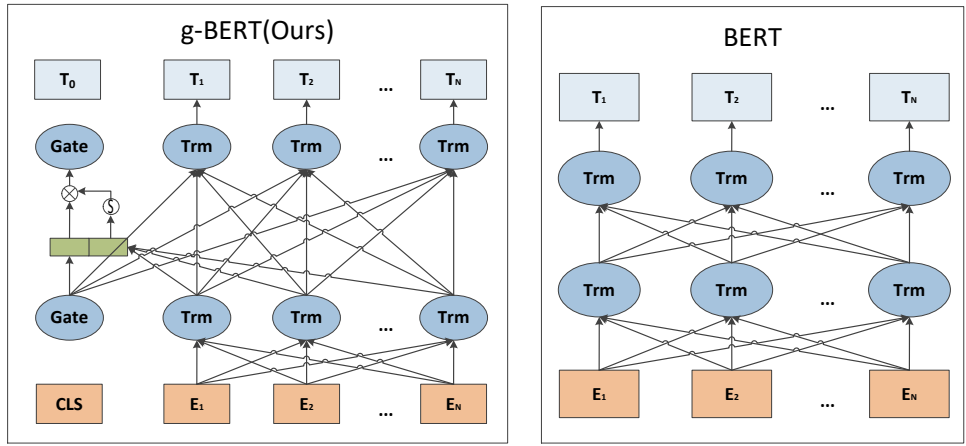

Fig.1 The difference between BERT and our model architecture

3.2 Pre-training Tasks

Our pre-training consists of three unsupervised prediction tasks: (1) masked language models like ERNIE [14]. (2) next sentence prediction like Devlin et al. [1]. So the above two parts are not repeated in this paper. (3) CWS and POS tagging.

Basic language tasks such as CWS and POS tagging help understand sentences. Since it costs too much when using Glod to label data, instead, we use Stanford NLP to label CWS and POS tagging in wikipedia. Although the labelling is not 100% accurate, this model can learn more information without supervision. Meanwhile, because of the existence of MLM and NSP, there is no need to worry that this model only learns the relevant information of CWS and POS tagging while ignores the irrelevant information of other CWS and POS tagging.

It is noted that POS tagging in Chinese is labeled on the basis of words, so we simply label the same POS tagging for all characters in the word. Combined with CWS labeling, we can get POS tagging at the word level. The overall structure is shown as Fig.2.

The training loss is the sum of the mean masked LM likelihood, mean next sentence prediction likelihood, CWS prediction likelihood and 0.5 POS tagging prediction likelihood.

The pre-training setting is consistent with that in Devlin et al. [1] Section 3.4.

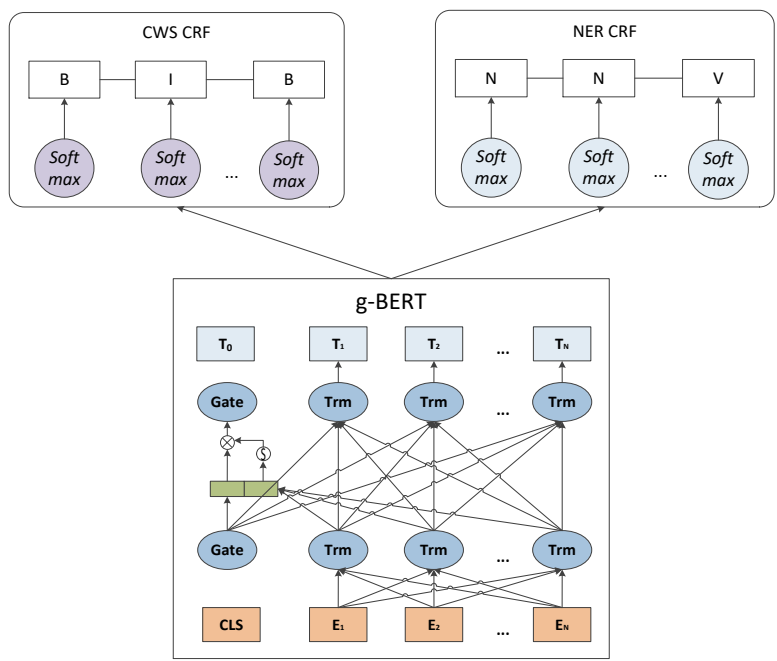

Fig.2 g-BERT model architecture

3.3 Pre-training procedure

In the pre-training procedure, we use wiki Chinese corpus and Stanford nlp to label. CWS is labeled as BIO and "O" for pad; POS tagging and parsing uses 65 categories, including (root, ip, pad). The same labeling is applied to each character in the same word, which is equivalent to Stanford nlp. The maximum length of a sentence in pre-training is 256. The weight of task loss such as CWS, POS tagging and parsing is 0.1, the weight of MLM and NSP task is 1.0 and the initial learning speed is 1e-4 with batch size 64.

In the process of words by MLM and NSP, the model can grasp non-local semantics. CWS, POS tagging and parsing can help the model be aware of the word boundary and sentence structure. Parsing determines the syntactic structure of the sentence or the dependency between words in a sentence, the result of which is generally presented in a tree data structure. In this paper, the tree data structure is transformed into two sequence labeling tasks. The first labeling sequence is POS tagging of the direct ancestor of the word. The second labeling sequence is POS tagging of the interval ancestor of the word, regardless of the location and information of the ancestral word. For example, in the sentence "句法分析是自然语言处理中关键技术之一" (Syntactic analysis is one of the key techniques in natural language processing), its labeling sequence is shown in Tab. 1 below. For the sake of simplicity, BIO annotation is used to label CWS and each character in the word is equally the same token. For instance, the corresponding CWS labeling to "分析" is "BI", POS tagging is "VV VV", and SP-part1 labeling is "VP" VP". Its boundary information can be obtained by combining analysis labels.

Tab.1 Parsing in tree structure

| Word | 句法 | 分析 | 是 | 自然 | 语言 | 处理 | 中 | 关键 | 技术 | 之一 |
|---|---|---|---|---|---|---|---|---|---|---|
| Translation | Syntactic | analysis | is | natural | language | processing | in | key | technology | one of |
| POS tagging | NN | VV | VC | AD | NN | NN | LC | JJ | NN | NN |
| SP-part1 | NN | VP | VP | ADVP | NP | NP | LCP | ADJP | NP | NP |
| SP-part2 | IP | IP | IP | IP | LCP | LCP | IP | NP | NP | NP |

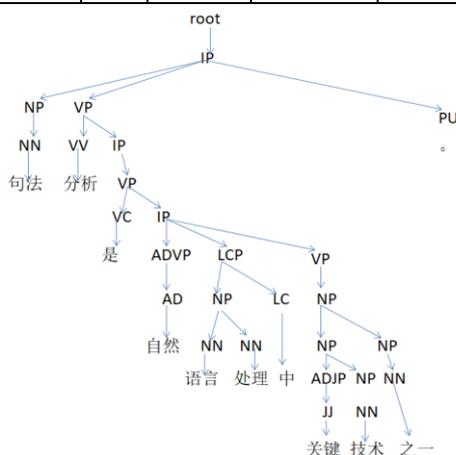

Fig.3 Parsing in tree structure

## 4 Expertiments

### 4.1 Datasets

In order to evaluate the model we proposed on Chinese NER, we conduct experiment on widely used dataset, including SIGHAN2006 NER dataset (Sighan-NER) [34]. The SighanNER is operated in simplified Chinese containing three entity types (person, location and organization). Since SighanNER does not include development set, we take 10% data in training set as sample in development set.

We take Precision(P), Recall(R) and F1 score as the evaluation criteria in our experiment. Hyper-parameter configurations are adjusted based on the performance of development set in Chinese NER task.

Initial learning rate is 0.00005, droupout rate is 0.1, weight decay is 0.01, batch size is 32, the length of the sequence is128.

4.2 Evaluation on SighanNER

Table 2 lists the comparison between SighanNER dataset and it is observed that the model we proposed achieves a state-of-the-art performance.

In the first block, we give the performance of previous methods for Chinese NER task in SighanNER dataset. Chen et al. [35] proposed a character-based CRF model for Chinese NER task.

Zhou et al. [36] introduced a pipeline model, which first segmented the text in character level CRF model and then applied word-level CRF to label. Cao et al. [13] extracted the public features of CWS and NER tasks, achieving F190.64%. F1 can be achieved as 92.6% when making fine tunes by using pre-trained bert-base model. ERNIE [14] replaced the random mask of single char in MLM with a word, making a 1.5% improvement with F1 reaching 93.8%. Compared with the state-of the-art model BERT-base, our method raises F1 score from 93.8% to 96.5%, while reducing 60% calculation quantity.

Table 2: Results on SighanNER dataset.

| Models | P(%) | F1(%) |
|---|---|---|
| Chen et al. (2006) | 91.22 | 86.20 |
| Cao et al.(2018) | 91.73 | 90.64 |
| BERT | 94.00 | 92.60 |
| ERNIE | 95.00 | 93.80 |
| g-BERT+POS tagging+pre-training | 96.6 | 96.5 |

4.3 Effectiveness of g-BERT and pre-training

Table 3 provides the experimental results of our proposed model and baseline as well as its simplified models on SighanNER dataset. The simplified models are described as follows:

- BERT+CRF: The $BERT_{BASE}$ model is used as strong baseline in our work, which is trained using Chinese NER training data. L=12, H=768, A=12, totle parameters=110M.
- g-BERT+CRF: We apply gating mechanism to BERT+CRF model. L=12, H=192, A=4,

totle parameters=40M.

- g-BERT+pre-training+CRF: We added the pre-training tasks of POS tagging, CWS and parsing. L=12, H=192, A=4, totle parameters=40M.

- Effectiveness of gating mechanism.

    g-BERT+CRF improves F1 score from 89.13% to 89.89% as compared with BERT+CRF on SighanNER dataset, and reduces computation by 60%

- Effectiveness of POS tagging, CWS, parser, pre-training.

By introducing adversarial training, g-BERT+pre-training+CRF boosts the performance as compared with g-BERT+CRF and BERT+CRF model, showing 0.15% and 0.36% improvement on SighanNER dataset. It proves that adversarial training can prevent specific features of CWS task from creeping into shared space.

Table 3: Comparison between our proposed model and simplified models on SighanNER dataset

| Models | P(%) | R(%) | F1(%) |
|---|---|---|---|
| BERT+CRF | 93.2 | 93.6 | 93.5 |
| g-BERT+CRF | 92 | 91.1 | 91.5 |
| g-BERT +pre-training+CRF | 96.6 | 96.5 | 96.5 |

## 5 Conclusions

In this paper, we first promote the model's understanding of sentence by pre-train task by adding CWS and POS tagging and use Stanford NLP tool to label the data. Second, the global node is added in the representation of sentence, the max-pooling of each sentence is input, the modified content of gate node is output and BERT model is improved, reducing calculation quantity by 60%. Finally, the model is evaluated in Chinese NER dataset. The experimental results show that the performance of our model improves by 3% with Test F1 to 96.5.